\documentclass[runningheads]{llncs}
\usepackage[T1]{fontenc}
\usepackage{hyperref}
\usepackage{booktabs}
\usepackage{makecell}
\usepackage{amsmath}
\usepackage{placeins}
\usepackage{booktabs}
\usepackage{makecell}
\usepackage{placeins}
\usepackage{graphicx}
\usepackage{amsmath}
\usepackage{amssymb}
\usepackage{url}
\usepackage{graphicx}
\begin{document}

\title{SGRNet: Spatially Guided Radiology Network for Structured Radiological Reporting of Head and Neck Cancer}
\titlerunning{SGRNet for Structured Reporting of HNC}

\author{Ayush Gupta\inst{1,2} \and Vinkle Srivastav\inst{1,2,3} \and Prateek Upadhya\inst{1,2} \and Amit Gupta\inst{4} \and Krithika Rangarajan\inst{4} \and Nicolas Padoy\inst{1,2}}
\authorrunning{A. Gupta et al.}

\institute{University of Strasbourg, CNRS, INSERM, ICube, Strasbourg, France \and IHU Strasbourg, France \and Department of Data Science and AI, Wadhwani School of Data Science and AI (WSAI), Indian Institute of Technology (IIT) Madras, Chennai, India  \and All India Institute of Medical Sciences, New Delhi, India\\ \email{ayush.gupta2@etu.unistra.fr}}

\maketitle
\vspace{-78mm}
\begin{center}
\textit{Accepted at the \href{https://caption-workshop.github.io/}{MICCAI-CaPTion 2026 Workshop}}
\end{center}
\vspace{68mm}

\begin{abstract}
Automated radiological report generation can alleviate clinical workloads and eliminate observer variability. However, standard free-text generation models pose hallucination risks in dense regions and fail under data scarcity. We address these challenges in Head and Neck Cancer (HNC) from contrast-enhanced CT (CECT) imaging. To enforce factual safety, we reformulate report generation as an anatomically grounded, multi-label, structured reporting task, predicting localized tumor involvement across a hierarchical clinical schema. To bridge the visual gap from missing metabolic imaging (e.g., PET), we introduce SGRNet (Spatially Guided Radiology Network), incorporating two low-cost spatial priors: automated organ segmentations and weakly supervised tumor localization maps modeled via 3D Gaussian heatmaps. These priors are dynamically integrated via spatial feature modulation to guide the network toward subtle tumor-induced structural alterations. Evaluated on a multi-centric dataset of 184 paired HNC CECT volumes and reports, on five clinically salient, densely packed anatomical subsites, SGRNet achieves a mean Average Precision (mAP) of 0.60, an 8.8 percentage-point absolute improvement over strong volume-only 3D baselines.

\end{abstract}

\section{Introduction}
Radiological reports underpin the clinical utility of medical imaging, yet manual reporting remains an operational bottleneck. This burden is acute in head and neck cancer (HNC), an anatomically dense disease site accounting for roughly 890,000 new cases annually \cite{iarc2022globocan}, where assessing tightly packed subsites (mucosal surfaces, deep spaces, neurovascular bundles) averages 24 minutes per scan \cite{macdonald2013radiologist_workload}, making it an ideal target for automation.

Hybrid modalities like PET/CT leverage metabolic tracers to highlight tumor boundaries but are cost-prohibitive. Contrast-enhanced computed tomography (CECT) remains the frontline standard of care but lacks direct metabolic signaling (see Appendix~\ref{sec:comp_three_images}), forcing tumor localization to rely on subtle tissue attenuation changes alone. Reporting models have progressed in 2D chest X-rays using benchmarks like CheXpert \cite{irvin2019chexpert} or 3D thoracic data like CT-RATE \cite{hamamci2024developing}, but fail to generalize to the intricate anatomy of the head and neck.

Furthermore, applying unconstrained autoregressive Vision-Language Models (VLMs) introduces severe safety risks where text hallucinations can radically alter staging. To ensure clinical safety, we reformulate report generation into an anatomically grounded structured reporting paradigm based on a hierarchical schema \cite{Gupta2025_EnhancedRadiologicalReportingHNC}, mapping imaging features to binary indicators of tumor involvement. To drive this domain despite extreme data scarcity, we curated a multi-centric dataset of 184 HNC CECT scans paired with radiologist-validated reports. To overcome low soft-tissue contrast, we propose SGRNet (Spatially Guided Radiology Network), which incorporates two low-cost spatial priors: automated organ segmentations and weakly supervised tumor localization maps modeled via 3D Gaussian heatmaps. These streams are fused via a spatial feature modulation mechanism that dynamically guides the network's attention toward subtle tumor-induced structural alterations. 

Our primary contributions are: (1) We introduce a specialized, multi-centric dataset of 184 paired HNC CECT volumes and structured reports. (2) We propose SGRNet, a prior-guided 3D architecture blending anatomical segmentation and weak pathological localization, providing a scalable alternative to voxel-level labeling. (3) On five clinically salient, densely packed anatomical subsites selected from the full structured schema (Appendix~\ref{sec:appendix_structured_temp}) as an initial, high-value validation of the approach, we demonstrate that our framework achieves an mAP of 0.60, marking an 8.8\% relative improvement over baseline 3D architectures, with high gains in small, complex structures.
\section{Related Work}
\textbf{Radiological Report Generation in 2D and 3D Domains:} Large-scale paired image–text benchmarks like MIMIC-CXR~\cite{johnson2019mimic} have fueled advanced 2D multi-modal models for free-text and controlled structured prediction~\cite{wu2024maira2}, but do not extend to 3D volumetric relationships. Volumetric 3D report generation models like CT2Rep~\cite{Ham_CT2Rep_MICCAI2024} and 3D-CT-GPT~\cite{chen2024_3d_ct_gpt} rely on thoracic benchmarks like CT-RATE~\cite{hamamci2024developing}. Their unconstrained text decoders introduce unacceptable hallucination risks if applied to highly intricate head and neck imaging workflows where sub-centimeter errors alter treatment trajectories.

\noindent\textbf{Head and Neck Imaging Benchmarks:} Publicly accessible HNC datasets are exceptionally rare and primarily center on metabolic PET/CT imaging~\cite{saeed2025multimodalHNC} and lack corresponding text reports. Standalone anatomical segmentation benchmarks exist (e.g., Walter et al. segmenting 71 distinct structures~\cite{Walter2024Segmentation}), but are bounded to pixel labeling. Our framework bridges accessible, standard-of-care contrast-enhanced CT (CECT) volumes with clinical reports, explicitly utilizing automated segmentation outputs like those from~\cite{Walter2024Segmentation} as architectural constraints rather than end-goals.

\noindent\textbf{Weakly Supervised Spatial Priors:} To bypass the steep costs of generating voxel-level 3D annotations, weak supervision has emerged as a key strategy. Methods leveraging bounding boxes or extreme points have successfully guided models with minimal manual overhead~\cite{roth2021goingextremes}. Prior works have adapted radiologists' eye-tracking gaze points to establish spatial focus~\cite{Zhong_MICCAI2024_Gaze}. Our framework uniquely repurposes 3D Gaussian tumor heatmaps as a low-cost, weak localization prior explicitly integrated to regularize and guide a multi-label structured reporting pipeline.

\section{Dataset and Curation Pipeline}
\textbf{Cohort Selection and Multi-Centric Heterogeneity:} We curated a multi-centric dataset comprising 184 contrast-enhanced CT (CECT) scans across two distinct clinical streams: (1) \textit{Public Domain Stream:} 49 scans extracted from the ACRIN-HNSCC study (ACRIN 6685)~\cite{Kinahan2019ACRIN6685_TCIA} hosted on TCIA, isolated after a quality audit to remove scans with severe motion or artifact profiles. (2) \textit{Institutional Clinical Stream:} 135 contrast-enhanced CT scans consecutively collected from the All India Institute of Medical Sciences (AIIMS), New Delhi, publicly available through the Indian Biological Image Archive (IBIA) under accession \texttt{CTS\_1000000034}\footnote{\url{https://ibdc.dbt.gov.in/ibia/study_details_browse_l/CTS_1000000034/}}. All institutional samples were de-identified and approved by local ethical review boards before downstream modeling. Pathologically, primary tumor involvement spans key sub-sites including the supraglottis, transglottis, nasopharynx, and oropharynx.

\noindent\textbf{Standardized Curation and Ground-Truth Structuring:} To establish reliable, clinically standardized labels from unstructured text, a dual-stage pipeline was enacted based on the clinical schema in~\cite{Gupta2025_EnhancedRadiologicalReportingHNC}: (1) For the AIIMS scans ($N=135$), native narrative free-text reports (average length: 275 words) were mapped to a structured template using an LLM to identify binary tumor involvement across subsites, followed by manual expert validation. (2) For the public ACRIN-HNSCC scans ($N=49$), which lacked matching text records, a senior radiologist manually evaluated each CECT volume and directly populated the identical structured template.

\section{Method}
We reformulate report generation into a structured multi-label classification task over an itemized anatomical schema to ensure factual safety. SGRNet processes raw CECT volumes alongside low-cost structural and pathological priors to regularize latent features.

\subsection{Clinically Controlled Schema Mapping}
Unstructured text reports are converted using the schema extraction prompt in~\cite{Gupta2025_EnhancedRadiologicalReportingHNC} with GPT-4. Unmentioned structural fields are designated as \textit{``Missing''}. Every output vector is audited by expert radiologists, serving as a reliable mapping intermediate between volumetric image patterns and structured categorical descriptors without generation artifacts.

\subsection{SGRNet Architectural Framework}
SGRNet employs a multi-branch architecture consisting of a primary imaging stream and two auxiliary prior streams that synthesize anatomical and pathological geometry.
\begin{figure}[t]
    \flushleft
    \includegraphics[page=1,width=1.0\linewidth]{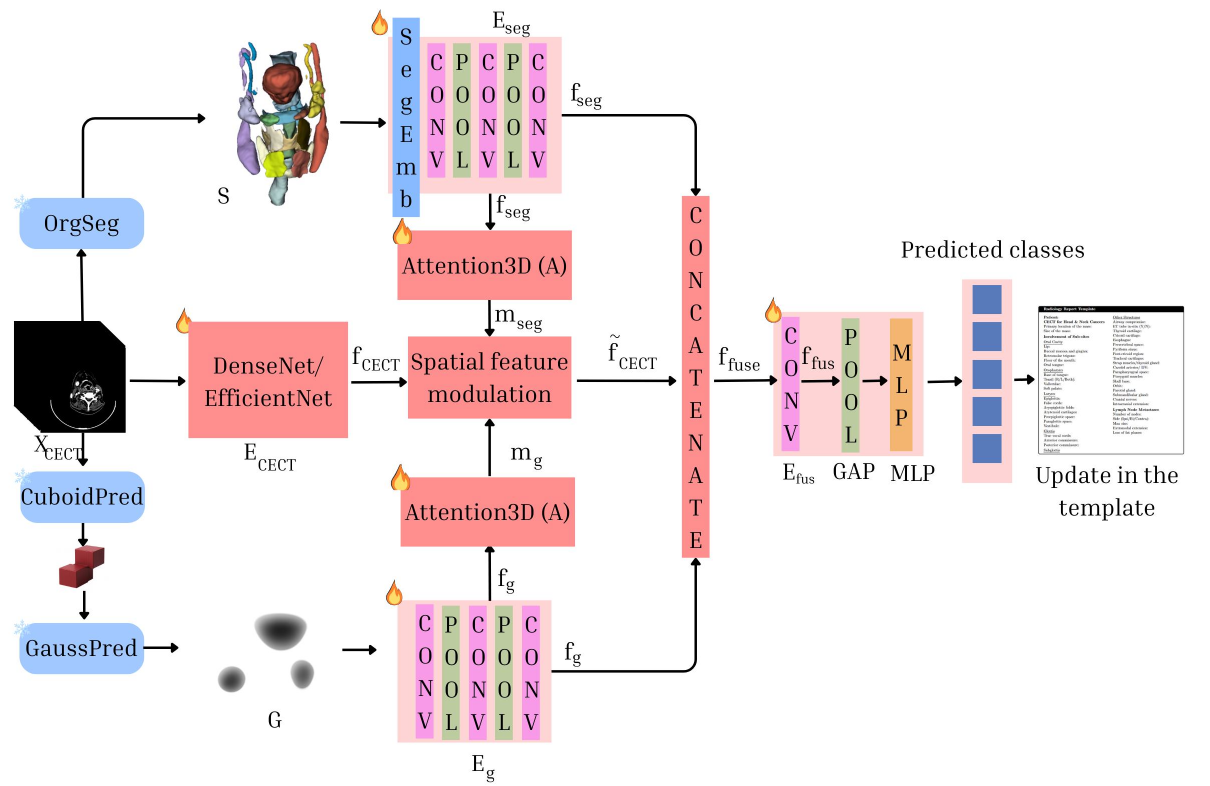}
    \caption{Architecture of the SGRNet. The tumour heatmap and the organ segmentation stream modulate the CECT stream via spatial feature modulation, allowing the network to emphasise tumour-salient regions while preserving global anatomical context.}
    \label{fig:attention_architecture}
\end{figure}

\subsubsection{Automated Structural Priors (OrgSeg):}
Anatomical context is derived using TotalSegmentator~\cite{Walter2024Segmentation} via a pre-trained nn-UNet engine~\cite{Isensee2021nnUNet} configured to extract 21 high-risk anatomical subsites relevant to HNC. The discrete categorical label map $S \in \{0,\dots,21\}^{D \times H \times W}$ is passed through a learned low-dimensional semantic embedding layer to save memory: $E_{\mathrm{emb}}(S) = \mathrm{Embed}(S) \in \mathbb{R}^{d_e \times D \times H \times W}$ (where $d_e = 8$). This dense map is passed into a 3D convolutional encoder ($E_{\mathrm{seg}}$) to produce downsampled structural prior feature maps:
\begin{equation}
f_{\mathrm{seg}} = E_{\mathrm{seg}}(E_{\mathrm{emb}}(S)) \in \mathbb{R}^{C_{\mathrm{seg}} \times d \times h \times w}
\end{equation}

\subsubsection{Weak Pathological Priors via 3D Gaussian Heatmaps:}
\label{subsec:gaussian_heatmap}
To model tumor-salient regions without requiring costly pixel-level segmentation masks, a 3D nn-UNet pre-trained on external CT channels from standard PET/CT data~\cite{saeed2025multimodalHNC} is fine-tuned on our 39 box-annotated training samples to output coarse binary cuboids. Let $\Omega \subset \mathbb{R}^3$ represent the connected foreground voxels of the thresholded bounding box. We extract its center of mass $(\mu_x, \mu_y, \mu_z)$ and determine its maximum spatial spans along the primary axes to yield bounding dimensions $(w_\mathrm{cuboid}, h_\mathrm{cuboid}, d_\mathrm{cuboid})$. A continuous 3D Gaussian heatmap $G$ is mathematically modeled across the volume space as:
\begin{equation}
G(x,y,z) = \exp\left(-\left( \frac{(x-\mu_x)^2}{2\sigma_x^2} + \frac{(y-\mu_y)^2}{2\sigma_y^2} + \frac{(z-\mu_z)^2}{2\sigma_z^2} \right)\right)
\end{equation}
where directional standard deviations $\sigma_x, \sigma_y, \sigma_z$ are set directly proportional to the bounding dimensions $w_\mathrm{cuboid}, h_\mathrm{cuboid}, d_\mathrm{cuboid}$, respectively. This continuous volume is mapped through a dedicated 3D Gaussian encoder $E_g$ to produce the pathological prior map: $f_g = E_g(G) \in \mathbb{R}^{C_G \times d \times h \times w}$.

\subsubsection{Spatial Feature Modulation:}
The prior feature maps are passed through a lightweight spatial gating block $A(\cdot)$ comprising $1\times1\times1$ convolutional layers and a sigmoid activation function to yield a unified spatial focus map $m = A(f_{\mathrm{seg}}) + A(f_g) \in \mathbb{R}^{1 \times d \times h \times w}$. Concurrently, the primary raw CECT volume $X_{\mathrm{CECT}}$ is processed by the primary visual trunk $E_{\mathrm{CECT}}$ to yield $f_{\mathrm{CECT}} \in \mathbb{R}^{C_{\mathrm{CECT}} \times d \times h \times w}$. Spatial feature modulation is executed via an element-wise soft-gating mechanism:
\begin{equation}
\tilde{f}_{\mathrm{CECT}} = f_{\mathrm{CECT}} \odot (1 + m)
\end{equation}
where $\odot$ denotes element-wise multiplication, amplifying subtle lesion signatures while preserving global anatomical alignment.

\subsubsection{Unified Feature Fusion and Structured Prediction:}
The modulated feature volume $\tilde{f}_{\mathrm{CECT}}$ is directly concatenated with the raw prior feature maps $f_{\mathrm{seg}}$ and $f_g$, passed through a 3D convolutional fusion network $E_{\mathrm{fus}}$, globally average pooled, and evaluated by an MLP to compute final categorical task probabilities:
\begin{equation}
\hat{y} = \sigma\Big(\mathrm{MLP}\big(\mathrm{GAP}(E_{\mathrm{fus}}(\mathrm{Concat}(\tilde{f}_{\mathrm{CECT}}, f_{\mathrm{seg}}, f_g))))\big)\Big)
\end{equation}
where $\hat{y} \in [0,1]^N$ represents the structured task vector mapped onto the $N$ core targets of the report template.

\section{Experiments \& Results}
\subsection{Implementation Details}
All network variants were evaluated using a stratified 5-fold cross-validation configuration on the core 145-scan modeling cohort across the five most consistently annotated anatomical subsites: \textit{Tongue}, \textit{Hypopharynx}, \textit{Larynx\_air}, \textit{Strap muscles + Thyroid gland}, and \textit{Carotid arteries + Internal jugular vein (IJV)}. The low-dimensional segmentation embedding block was configured with $d_e = 8$, yielding downsampled feature representations of size $42 \times 32 \times 32$ with 64 channels across all auxiliary trunks. To prevent overfitting, standard 3D spatial transformations (random scaling, rotation, and elastic deformation) were enforced during optimization. Models were trained end-to-end utilizing the Adam optimizer with a learning rate of $1\times10^{-5}$ and a batch size of 2. Severe clinical class imbalances were mitigated using the Asymmetric Loss (ASL) formulation:
\begin{equation}
\mathcal{L}_{\text{ASL}} = - \sum_{c=1}^{K} \Big[ y_c \,(1 - p_c)^{\gamma_+} \, \log(p_c) + (1 - y_c) \, p_c^{\gamma_-} \, \log(1 - p_c) \Big]
\end{equation}
where $\gamma_+ = 1$ and $\gamma_- = 2$ were fixed empirically. Task execution is reported via class-specific and mean Average Precision (mAP). For the weak pathological stream, the localized 3D nn-UNet cuboid engine was pre-trained on the external CT channels of a public PET/CT cohort~\cite{saeed2025multimodalHNC} and fine-tuned for 250 epochs over the 39 box-annotated volumes using a cosine annealing learning rate scheduler, achieving a mean Dice score of 0.51 compared to 0.37 when optimized from scratch.

\subsection{Quantitative Results and Ablation Analysis}
The upper portion of Table~\ref{tab:results_tab} presents a performance evaluation comparing volume-only 3D backbones against our prior-guided variants. Baseline networks relying purely on unconstrained CECT inputs demonstrated moderate predictive capabilities, reaching mAPs of 0.512 (DenseNet121) and 0.494 (EfficientNet-B0). The 3D foundation model (CT-FM) achieved competitive metrics on macro-anatomical pathways but deteriorated sharply on intricate soft-tissue subsets like the strap muscles (0.256 AP). 

We additionally report a non-learned \textit{Simple Overlap Baseline}, which flags a structured field as involved whenever the fractional spatial overlap between the corresponding TotalSegmentator organ mask and the fine-tuned tumor mask exceeds a fixed threshold, with no learned classification component; this isolates how much of the observed gains stem from simply possessing the two spatial priors versus their learned, end-to-end fusion in SGRNet.

By contrast, SGRNet variants achieved substantial, uniform improvements across nearly all highly packed targets. SGRNet instantiated on a DenseNet121 trunk achieved an optimal mAP of \textbf{0.600}, registering a notable 8.8\% relative improvement over its corresponding raw image baseline. These margins are particularly pronounced in ultra-dense structures like the hypopharynx (+0.151 AP) and the laryngeal airway (+0.184 AP).

The lower portion of Table~\ref{tab:results_tab} presents a detailed ablation matrix parsing the contributions of individual spatial priors and fusion architectures. The empirical results show that spatial modulation mechanisms consistently outperform simple element-wise addition across both backbones. Integrating anatomical organ maps alone provides a steady baseline boost, but the subsequent integration of the weak Gaussian tumor heatmap acts as a critical regularizer, driving the DenseNet mAP to its peak performance tier of 0.600.

\begin{table*}[!ht]
\centering
\caption{Average Precision (AP) for baseline CECT models, the non-learned overlap baseline, prior-blending ablations, and the final SGRNet variants. (IJV: Internal Jugular Vein. Notable values highlighted in bold; final SGRNet rows shown in bold).}
\label{tab:results_tab}
\scriptsize 
\setlength{\tabcolsep}{2.5pt} 
\renewcommand{\arraystretch}{0.9} 
\begin{tabular}{llcccccc}
\toprule
\textbf{Model} & \textbf{Input} & \textbf{Tongue} & \textbf{\makecell{Hypo-\\pharynx}} & \textbf{\makecell{Larynx\\Air}} & \textbf{\makecell{Strap +\\Thyroid}} & \textbf{\makecell{Carotid\\+ IJV}} & \textbf{\makecell{Mean\\AP}} \\
\midrule
DenseNet121 & CECT Only & \textbf{0.549} & 0.559 & 0.574 & 0.357 & 0.519 & 0.512 \\
EfficientNet-B0 & CECT Only & 0.543 & 0.534 & 0.563 & 0.339 & 0.489 & 0.494 \\
CT-FM~\cite{pai2025vision} & CECT Only & 0.419 & 0.576 & 0.612 & 0.256 & \textbf{0.579} & 0.488 \\
\midrule
Simple Overlap Baseline & Org+Tum-Overlap & 0.531 & 0.690 & 0.724 & 0.401 & 0.523 & 0.574 \\
\midrule
DenseNet121 & Org-Add & 0.504 & \textbf{0.725} & 0.699 & 0.314 & \textbf{0.627} & 0.574 \\
DenseNet121 & Org-Modul & 0.494 & 0.674 & \textbf{0.756} & 0.447 & 0.601 & 0.594 \\
EfficientNet-B0 & Org-Add & 0.475 & 0.708 & 0.658 & 0.283 & 0.593 & 0.543 \\
EfficientNet-B0 & Org-Modul & 0.478 & 0.636 & 0.701 & 0.413 & 0.600 & 0.567 \\
\midrule
DenseNet121 & Tum-Add & 0.528 & 0.688 & 0.701 & 0.342 & 0.588 & 0.569 \\
DenseNet121 & Tum-Modul & \textbf{0.536} & 0.702 & 0.714 & 0.355 & 0.596 & 0.581 \\
EfficientNet-B0 & Tum-Add & 0.512 & 0.671 & 0.682 & 0.328 & 0.571 & 0.553 \\
EfficientNet-B0 & Tum-Modul & 0.521 & 0.684 & 0.695 & 0.336 & 0.579 & 0.563 \\
\midrule
DenseNet121 & Org+Tum-Add & 0.485 & 0.731 & 0.715 & 0.367 & 0.614 & 0.582 \\
EfficientNet-B0 & Org+Tum-Add & 0.431 & 0.713 & 0.686 & 0.345 & 0.543 & 0.544 \\
\midrule
\textbf{SGRNet (DenseNet)} & Org+Tum-Modul & 0.478 & \textbf{0.710} & \textbf{0.758} & \textbf{0.501} & 0.554 & \textbf{0.600} \\
\textbf{SGRNet (Efficient)} & Org+Tum-Modul & 0.498 & 0.674 & 0.724 & 0.467 & 0.558 & 0.584 \\
\bottomrule
\end{tabular}
\end{table*}

\section{Discussion \& Limitations}
The experimental profiles confirm that injecting explicit spatial boundaries and weak pathological context optimizes multi-label classification accuracy within low-contrast 3D medical datasets. Raw, unconstrained volumes fail to provide standard 3D convolutional trunks or large foundation models (CT-FM) with the necessary fine-grained visual attention to parse small, packed structures like the strap muscles or larynx. While CT-FM utilizes a wide pre-training corpus, its representations remain generalized; our localized spatial feature modulation layer acts as a target-specific corrective gate, steering feature maps back to tumor-adjacent boundaries.

On the \textit{Tongue} subsite, the baseline DenseNet121 (0.549 AP) slightly outperformed our prior-guided framework (0.478 AP), likely because oral tongue tissue is highly mobile and deformable during CECT acquisition -- unlike rigid structures such as the carotid sheath or laryngeal cartilage, its cross-sectional boundary shifts with deglutition and jaw position, introducing misalignment noise into the TotalSegmentator-derived gating stream. For stable structures, however, the joint prior strategy delivers substantial margins, and the underlying pipeline is broadly scalable: external PET/CT datasets can be stripped to their CT channels to pre-train weak tumor localizers, reducing reliance on costly hybrid scanners.

\noindent\textbf{Limitations:} Despite its strengths, several limitations constrain our current pipeline. First, the core modeling size (145 cases) leaves the network sensitive to rare, highly atypical tumor morphologies. Second, modeling weak tumor regions as rigid axis-aligned 3D cuboids occasionally includes healthy adjacent boundaries within the computed Gaussian distributions. Third, errors generated by automated segmentation tools like TotalSegmentator can cascade into downstream feature modulation steps. Fourth, our evaluation targets five high-yield, densely packed subsites rather than the complete structured template (Appendix~\ref{sec:appendix_structured_temp}), chosen for annotation consistency and diagnostic complexity; we treat this as an initial validation of the reformulation, with full-schema scaling left to future work. Finally, we omit a head-to-head comparison against free-text/VLM report generators (e.g.,~\cite{Ham_CT2Rep_MICCAI2024,chen2024_3d_ct_gpt}), as they operate in an unconstrained text space with no shared metric or hallucination benchmark for HNC CECT; building such a benchmark is left to future work.

\section{Conclusion}
We have presented SGRNet, a prior-guided 3D architecture engineered for the safe, structured radiological reporting of head and neck cancer on contrast-enhanced CT scans. By converting text reports into deterministic, structured subsite targets, we eliminate the severe text hallucination risks tied to unconstrained vision-language models. By dynamically modulating visual features with low-cost anatomical segmentations and weak 3D Gaussian heatmaps, SGRNet successfully circumvents the lack of metabolic signals in standard imaging, yielding an 8.8\% relative mAP improvement over raw 3D baselines. Future expansions will focus on multi-institutional validation, semi-supervised representation learning to utilize unlabeled volumes, and investigating deformable registration steps to handle highly mobile oral structures like the tongue.

\section{Acknowledgements}
This work was partially supported by the Interdisciplinary Thematic Institute HealthTech, as part of the ITI 2021-2028 program of the University of Strasbourg, CNRS, and Inserm, funded by IdEx Unistra (ANR-10-IDEX-0002) and SFRI (STRAT’US project, ANR-20-SFRI-0012) under the framework of the French Investments for the Future Program. This work was also partially supported by French state funds managed by the ANR under Grant ANR-10-IAHU-02. This work was granted access to the HPC resources of IDRIS under the allocations AD011011631R4 made by GENCI. The authors would like to acknowledge the High-Performance Computing Center of the University of Strasbourg for supporting this work by providing scientific support and access to computing resources. Part of the computing resources was funded by the Equipex Equip@Meso project (Programme Investissements d'Avenir) and the CPER Alsacalcul/Big Data.

\newpage

%
%
%
%


\appendix

\renewcommand{\thefigure}{A.\arabic{figure}}
\setcounter{figure}{0}

\renewcommand{\thetable}{A.\arabic{table}}
\setcounter{table}{0}

\section{Comparison of PET/CT, PET/MRI and CECT}
\label{sec:comp_three_images}

\begin{figure}[!htbp]
\centering
\includegraphics[page=2,width=0.85\linewidth]{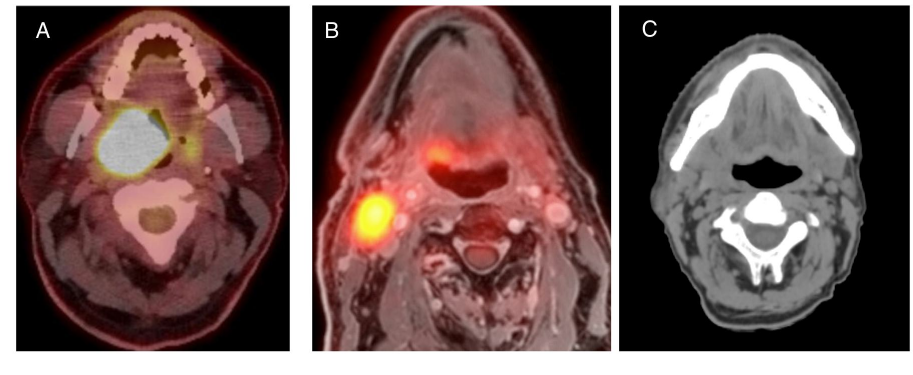}
\caption{
Visual comparison of imaging modalities.
(A) PET/CT highlights metabolically active tumour regions via radiotracer uptake.
(B) PET/MRI provides high soft-tissue contrast with PET-guided localisation.
(C) CECT lacks metabolic tracers, resulting in more subtle tumour boundaries and motivating the need for automated interpretability methods.
}
\label{fig:modality_comparison}
\end{figure}

\FloatBarrier
\section{Example of a Free-Text Radiology Report}
\label{sec:free_text_report}

\begin{quote}
\small

\textbf{Patient 1}

\medskip

\textbf{Procedure:} CECT of Face \& Neck.

\textbf{Clinical background:} Ca oropharynx (cT4aN3bM0) post CT/RT.

There is ill-defined enhancing soft tissue thickening in the oropharynx
measuring approximately 2.7 cm in longest dimension involving the right
lateral and posterior oropharyngeal wall, right vallecula, median
glossoepiglottic fold, the tip of the epiglottis, and right base of tongue.
Non-enhancing oedematous thickening of the bilateral aryepiglottic folds and
vocal cords is noted, likely representing post-radiotherapy changes.
The hyoid bone is normal. The prevertebral fat space is maintained.

Enlarged multiple conglomerated necrotic lymph nodes are present in bilateral
level II and III regions, the largest measuring approximately
2.0 $\times$ 2.4 cm on the right side. There is encasement of vessels on the
right side with thrombosis of the internal jugular vein extending from the
jugular foramen to the sigmoid sinus.

\textbf{Paranasal sinuses/nasal cavity:}
Mucosal thickening is seen in the right maxillary sinus.

\textbf{Maxilla/mandible:}
Normal.

\textbf{Infra-temporal neck spaces:}
Normal.

\textbf{Additional information:}
None.

\textbf{Comparison:}
Compared with the scan dated 17 June 2022, there is a reduction in the size
of the primary mass and lymph nodes.

\textbf{Impression:}
Irregular enhancing thickening involving the oropharynx with extensions as
described above and bilateral cervical conglomerated necrotic lymphadenopathy.

\end{quote}

\FloatBarrier
\section{Structured Reporting Template for Head and Neck Cancers}
\label{sec:appendix_structured_temp}

\begin{table}[!htbp]
\caption{Structured reporting template used for head and neck cancer CT examinations.}
\label{tab:structured_template}
\centering
\small

\begin{tabular}{p{0.47\textwidth} p{0.47\textwidth}}
\toprule

\textbf{General Information}
&
\textbf{Additional Structures}
\\

Patient ID
&
Airway compromise
\\

Primary tumour location
&
Endotracheal tube in-situ
\\

Tumour size
&
Thyroid cartilage
\\

&
Cricoid cartilage
\\

\midrule

\textbf{Oral Cavity}
&
Esophagus
\\

Lip
&
Prevertebral space
\\

Buccal mucosa and gingiva
&
Pyriform sinus
\\

Retromolar trigone
&
Post-cricoid region
\\

Floor of mouth
&
Tracheal cartilages
\\

Oral tongue
&
Strap muscles / thyroid gland
\\

\midrule

\textbf{Oropharynx}
&
Carotid arteries / IJV
\\

Base of tongue
&
Parapharyngeal space
\\

Tonsil (R/L/Both)
&
Pterygoid muscles
\\

Valleculae
&
Skull base
\\

Soft palate
&
Orbit
\\

&
Parotid gland
\\

\midrule

\textbf{Larynx}
&
Submandibular gland
\\

Epiglottis
&
Cranial nerves
\\

False cords
&
Intracranial extension
\\

Aryepiglottic folds
&
\\

Arytenoid cartilages
&
\textbf{Lymph Node Metastases}
\\

Preepiglottic space
&
Number of nodes
\\

Paraglottic space
&
Laterality
\\

Vestibule
&
Maximum size
\\

True vocal cords
&
Extranodal extension
\\

Anterior commissure
&
Loss of fat planes
\\

Posterior commissure
&
\\

Subglottis
&
\\

\bottomrule
\end{tabular}
\end{table}

\FloatBarrier
\section{Summary of Dataset Characteristics}
\label{sec:dataset_characteristics}

\begin{table}[!htbp]
\caption{Summary of dataset-level imaging characteristics.}
\label{tab:dataset_stats}
\centering
\small
\begin{tabular}{lcc}
\toprule
\textbf{Parameter} & \textbf{Value} & \textbf{Unit} \\
\midrule
Total scans & 184 & -- \\
Average voxel spacing & $1 \times 1 \times 1$ & mm$^{3}$ \\
Average resolution & $512 \times 510$ & pixels \\
Average number of slices & 288 & slices \\
Average mean HU & $-763$ (Std. 576) & HU \\
HU range & $[-1732,\;3444]$ & HU \\
\bottomrule
\end{tabular}
\end{table}

\FloatBarrier
\section{Organ Labels Used for Anatomical Segmentation}
\label{sec:org_label_map}

\begin{table}[!htbp]
\caption{Segmented organs and corresponding label identifiers.}
\label{tab:organ_labels}
\centering
\small
\begin{tabular}{cl}
\toprule
\textbf{ID} & \textbf{Structure} \\
\midrule
0 & Background \\
1 & Larynx air \\
2 & Thyroid cartilage \\
3 & Cricoid cartilage \\
4 & Hyoid bone \\
5 & Tongue \\
6 & Digastric (left) \\
7 & Digastric (right) \\
8 & Sternothyroid (left) \\
9 & Sternothyroid (right) \\
10 & Thyrohyoid (left) \\
11 & Thyrohyoid (right) \\
12 & Submandibular gland (left) \\
13 & Submandibular gland (right) \\
14 & Thyroid gland \\
15 & Internal carotid artery (left) \\
16 & Internal carotid artery (right) \\
17 & Internal jugular vein (left) \\
18 & Internal jugular vein (right) \\
19 & Trachea \\
20 & Oropharynx \\
21 & Hypopharynx \\
\bottomrule
\end{tabular}
\end{table}

\FloatBarrier
\section{Segmentation Results}
\label{sec:output_nn_unet_model}

\begin{figure}[!htbp]
\centering
\includegraphics[page=3,width=0.85\linewidth]{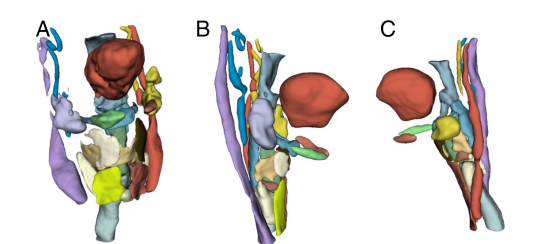}
\caption{
Qualitative results from the anatomical organ segmentation model.
Panels (A)--(C) illustrate representative outputs from three different
patients showing the predicted masks for the 21 anatomical structures.
}
\label{fig:org_seg_results}
\end{figure}

\begin{figure}[!htbp]
\centering
\includegraphics[page=4,width=0.85\linewidth]{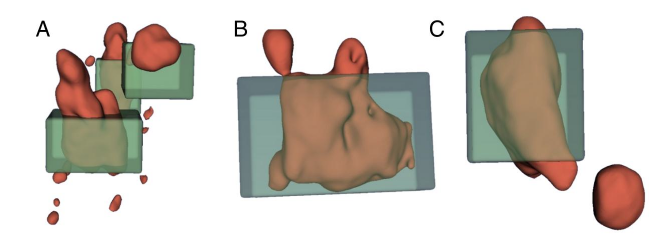}
\caption{
Tumour segmentation results from an nnU-Net model trained exclusively on
CT images from the public PET/CT dataset.
Panels (A)--(C) show representative examples.
}
\label{fig:tum_seg_direct}
\end{figure}

\begin{figure}[!htbp]
\centering
\includegraphics[page=5,width=0.85\linewidth]{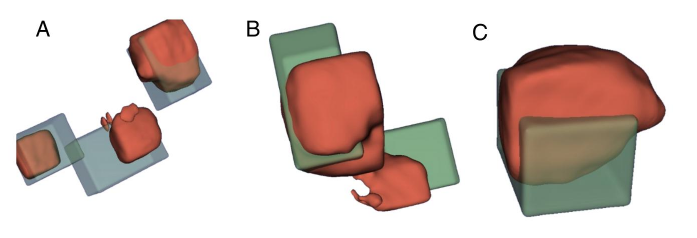}
\caption{
Tumour segmentation results from the nnU-Net model pre-trained on the
public PET/CT dataset and subsequently fine-tuned using the manually
curated cuboid regions.
Panels (A)--(C) show representative examples.
}
\label{fig:tum_seg_cuboid}
\end{figure}

\FloatBarrier
\section{Evaluation Metric}
\label{sec:equations}

\textbf{Dice Similarity Coefficient.}
The Dice similarity coefficient between two binary masks
$A$ and $B$ is defined as

\[
\mathrm{Dice}(A,B)
=
\frac{2|A \cap B|}
{|A| + |B|},
\]

where $A$ and $B$ denote the predicted and ground-truth masks,
respectively.


\end{document}